\documentclass[letterpaper, 10 pt, conference]{ieeeconf}  
\IEEEoverridecommandlockouts                              
\usepackage{amsmath,graphicx}
\usepackage{tikz}
\usepackage{comment}
\usepackage{amssymb} 
\usepackage{color}
\usepackage{multirow}

\usepackage{subcaption}
\usepackage{hyperref}
\usepackage{cleveref}
\usepackage{gensymb}

\usepackage[accsupp]{axessibility}  

\title{\LARGE \bf Challenges of Indoor SLAM: A multi-modal multi-floor dataset for SLAM evaluation}

\author{Pushyami Kaveti, Aniket Gupta, Dennis Giaya, Madeline Karp,\\ Colin Keil, Jagatpreet Nir, Zhiyong Zhang, Hanumant Singh${\ast}$%
\thanks{${\ast}$All authors are affiliated with Northeastern University, Boston. MA, USA}
}

\begin{document}

\maketitle
\thispagestyle{empty}
\pagestyle{empty}
\begin{abstract}
Robustness in Simultaneous Localization and Mapping (SLAM) remains one of the key challenges for the real-world deployment of autonomous systems. SLAM research has seen significant progress in the last two and a half decades, yet many state-of-the-art (SOTA) algorithms still struggle to perform reliably in real-world environments. There is a general consensus in the research community that we need challenging real-world scenarios which bring out different failure modes in sensing modalities. In this paper, we present a novel multi-modal indoor SLAM dataset covering challenging common scenarios that a robot will encounter and should be robust to. Our data was collected with a mobile robotics platform across multiple floors at Northeastern University's ISEC building. Such a multi-floor sequence is typical of commercial office spaces characterized by symmetry across floors and, thus, is prone to perceptual aliasing due to similar floor layouts. The sensor suite comprises seven global shutter cameras, a high-grade MEMS inertial measurement unit (IMU), a ZED stereo camera, and a 128-channel high-resolution lidar. Along with the dataset, we benchmark several SLAM algorithms and highlight the problems faced during the runs, such as perceptual aliasing, visual degradation, and trajectory drift. The benchmarking results indicate that parts of the dataset work well with some algorithms, while other data sections are challenging for even the best SOTA algorithms. The dataset is available at \url{https://github.com/neufieldrobotics/NUFR-M3F}. 
\end{abstract} 

\begin{figure}[t!]
\centering
\captionsetup{font={footnotesize}}


\begin{tabular}{c}
\subfloat[]{\includegraphics[width =0.45\columnwidth]{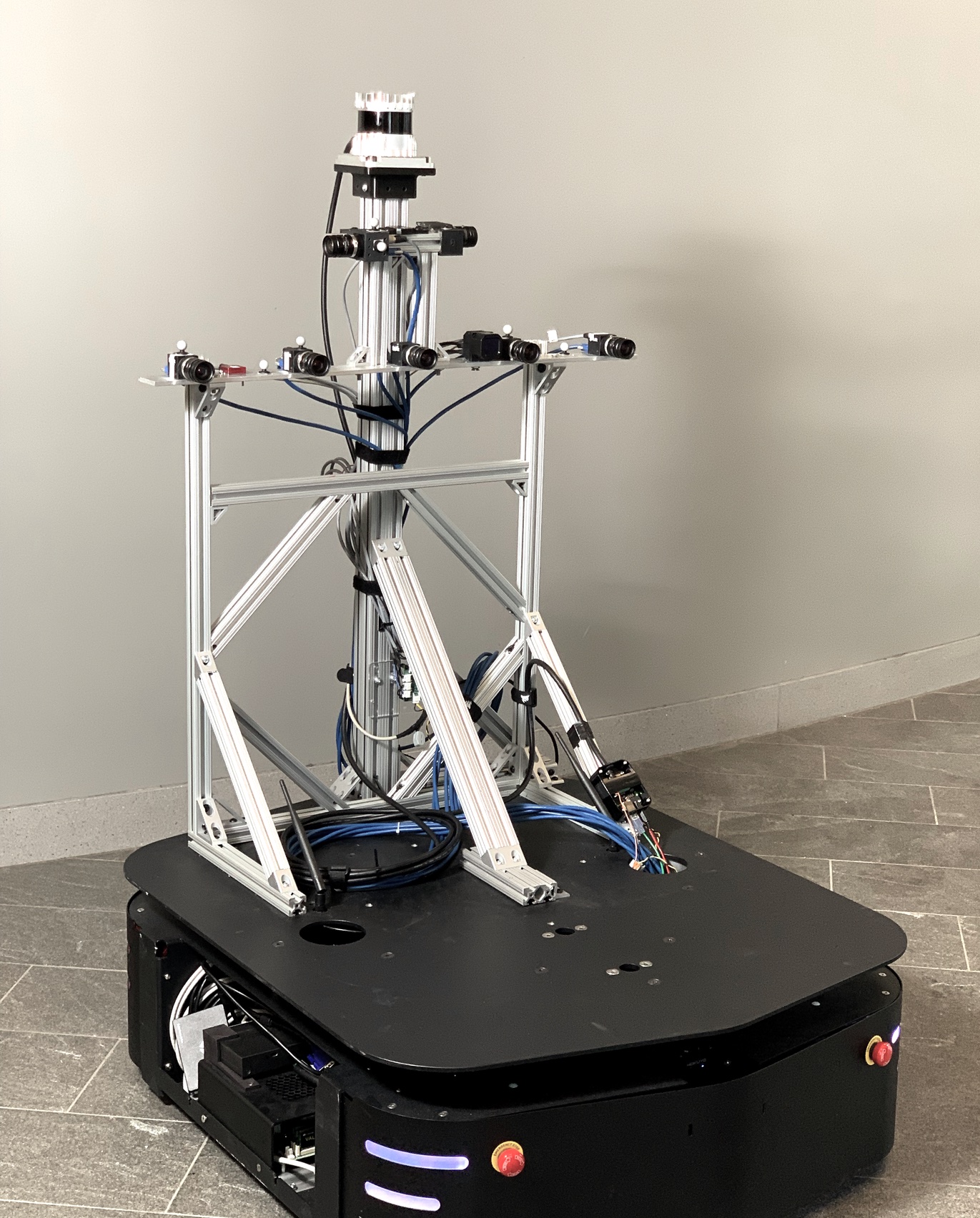}
\label{fig:system}}
\subfloat[]{\includegraphics[width =0.42\columnwidth]{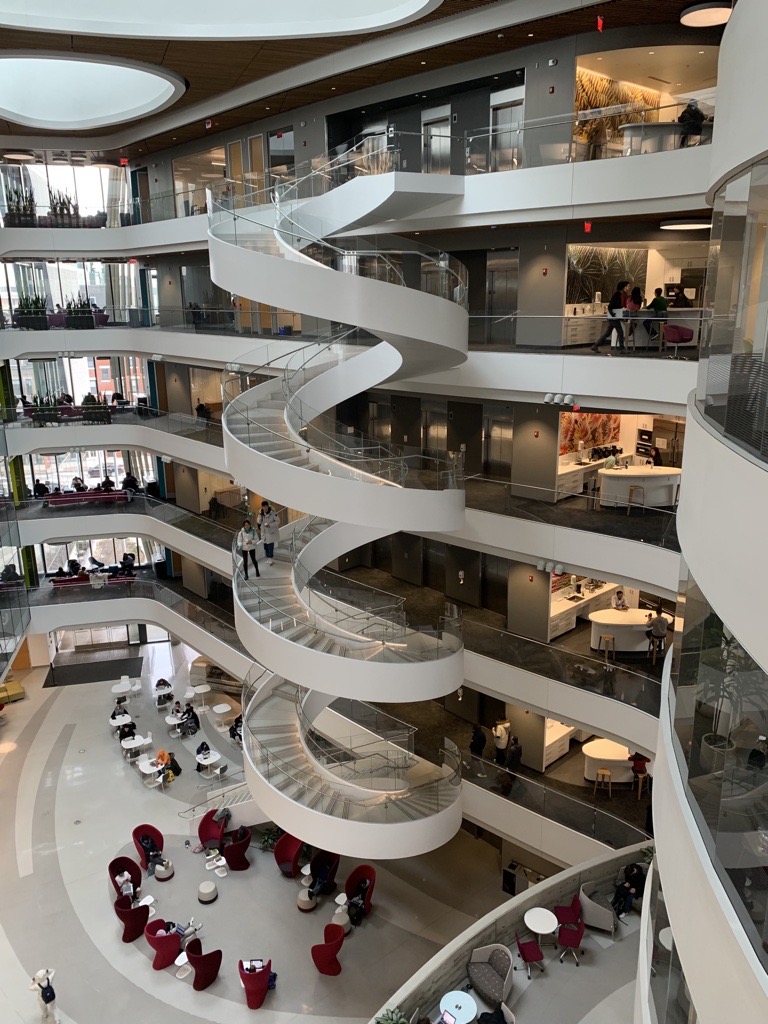}}\\
\subfloat[]{\includegraphics[width =0.84\columnwidth]{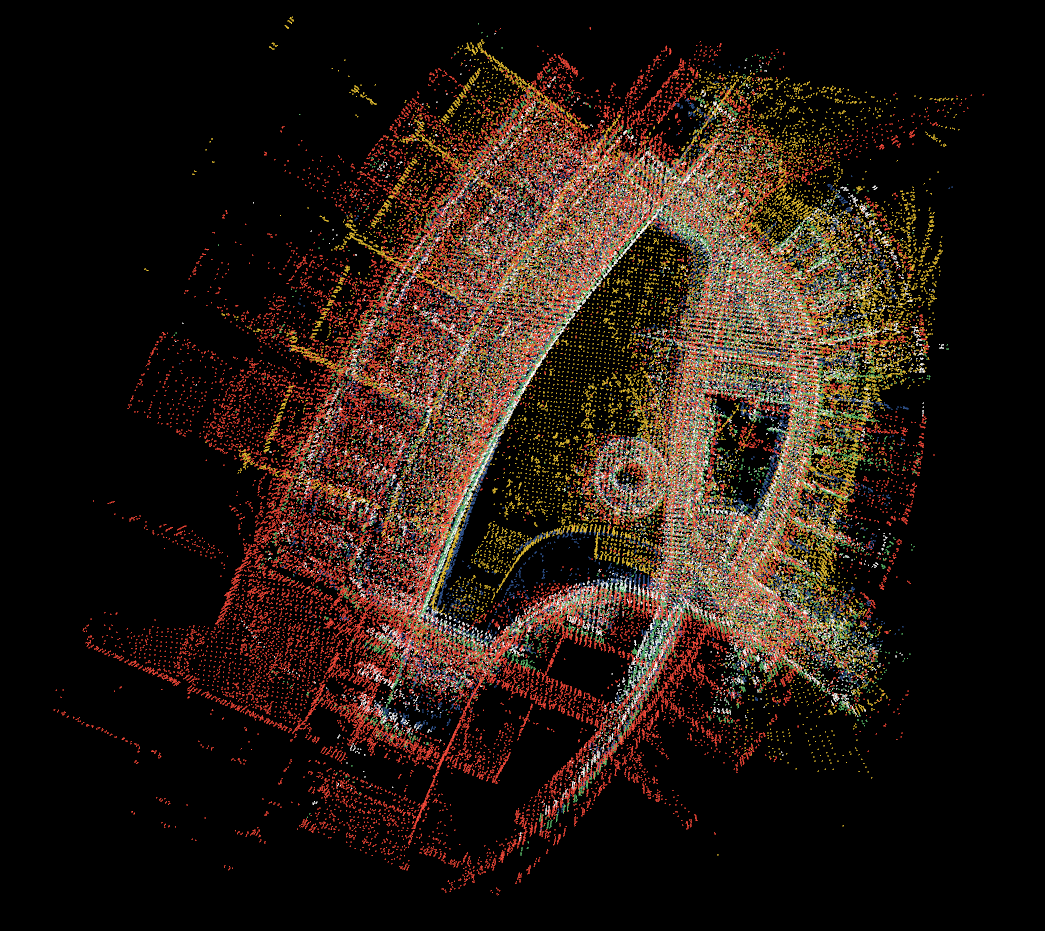}}
\end{tabular}
\caption{(a) The data collection rig mounts to an omnidirectional base with the sensors approximately 1.2m above the ground. (b) The data collection site, Northeastern University's Interdisciplinary Science and Engineering Complex (ISEC), which has an open atrium and several floors with a high degree of symmetry in their layout and overall design.(c) A composite rendition of the lidar point cloud depicting all the floors from top view.}
\label{fig:overview}
\end{figure}


\section{Introduction} 
This paper presents a multi-modal SLAM dataset of several real-world sequences captured in a large-scale indoor environment. Simultaneous Localization and Mapping is an extensively researched topic in robotics that has seen major advances in recent decades \cite{cadena}.  It is a hardware and software co-design problem, and the performance of the solution is a function of the right choice of complementary sensors, their proper configuration and calibration, vehicle motions, and, finally, the uncertainties in the real-world mapping environment. Often, methods that work well in certain scenarios fail in the real world due to various factors. These may include environmental uncertainties, dynamic objects, illumination artifacts, and issues associated with robotic motion and trajectories. 

The performance of state-of-the-art (SOTA) SLAM algorithms is limited by the lack of publicly available datasets for testing:  KITTI\cite{geiger2013vision}, TUM RGB-D\cite{sturm2012evaluating}, TUM Mono\cite{engel2016photometrically}, and  Euroc MAV\cite{burri2016euroc}.  These datasets have many strengths and have positively impacted the SLAM algorithm design and evaluation using different sensor modalities, including monocular vision, stereo vision, visual-inertial odometry (VIO), RGBD cameras, and 3D lidars. However, new large-scale public datasets with multiple sensing modalities are essential. Recent work \cite{zhang2021balancing}\cite{chghaf2022camera} has also shown that fusing data from multiple sensors improves the robustness and accuracy of SLAM estimates in challenging scenarios often encountered in the real-world.
\begin{table*}[t!]
\caption{An overview of the indoor multi-modal SLAM datasets}
\resizebox{\textwidth}{!}{%
\begin{tabular}
{|p{1.5cm}|p{2.2cm}|p{2cm}|p{1cm}|p{1cm}|p{1.7cm}|p{2cm}|p{1cm}|p{1.5cm}|}
\hline
\textbf{} &
  \textbf{Sensors} &
  \textbf{Arrangement} &
  \textbf{Frame Rate(Hz)} &
  \textbf{No. of Sequences} &
  \textbf{Platform} &
  \textbf{Ground truth} &
  \textbf{Sync} &
  \textbf{Environment} 
  \\
  \hline
\textbf{Newer College} &
  \begin{tabular}[c]{@{}l@{}}2 GS cameras\\ 1 LiDAR\\ 1 IMU\end{tabular} &
  2 stereo,  2 Non-overlapping &
  30 &
  3 &
  Handheld &
  \begin{tabular}[c]{@{}l@{}}Survey grade\\ 3D Imaging Laser\end{tabular} &
  HW + SW &
  Campus 
   \\
   \hline
\textbf{PennCosy-VIO} &
  \begin{tabular}[c]{@{}l@{}}3 RGB GS cameras\\ 2 Gray RS cameras\\ 2 IMUs\end{tabular} &
  3 min overlapping,  2 stereo &
  20 &
  4 &
  Handheld &
  \begin{tabular}[c]{@{}l@{}}Fiducial \\ Markers\end{tabular} &
  HW + SW &
  Campus 
   \\ 
   \hline
\textbf{HILTI} &
  \begin{tabular}[c]{@{}l@{}}5 GS Gray cameras\\ 2 LiDARS\\ 3 IMU\end{tabular} &
  2 stereo, 3 Non-overlapping & 
  10  &
  12 &
  Handheld &
  MoCAP &
  HW +SW &
  \begin{tabular}[c]{@{}l@{}}Construction \\ site\end{tabular} 
   \\ \hline
\textbf{HILTI-Oxford} &
  \begin{tabular}[c]{@{}l@{}}5 GS IR cameras\\ 1 LiDAR\\ 1 IMU\end{tabular} &
  2 stereo, 3 Non-overlapping &
  40  &
  16 &
  Handheld &
  \begin{tabular}[c]{@{}l@{}}Survey grade\\ 3D Imaging Laser\end{tabular} &
  HW + SW &
  \begin{tabular}[c]{@{}l@{}}Construction \\ Site\end{tabular} 
   \\ \hline
\textcolor{blue}{\textbf{Ours}} &
  \textcolor{blue}{\begin{tabular}[c]{@{}l@{}}7 GS RGB cameras\\ 1 Zed 2i \\ 1 LiDAR\\ 1 IMU\end{tabular}} &
  \textcolor{blue}{\begin{tabular}[c]{@{}l@{}}5 Fronto-Parallel\\ 2 Side-ways\end{tabular}} &
   \textcolor{blue}{20} &
  \textcolor{blue}{14} &
  \textcolor{blue}{Mobile Robot} &
  \textcolor{blue}{\begin{tabular}[c]{@{}l@{}}Fiducial markers\\ +3D \\LiDAR alignment\end{tabular}} &\textcolor{blue}{
  HW + SW} &
  \textcolor{blue}{\begin{tabular}[c]{@{}l@{}}Campus \\ \\ office\end{tabular}}
   \\ \hline
\end{tabular}
HW +
SW}
\label{tab:dataset_sensors}
\end{table*}

\begin{figure}[t!]
\centering
\captionsetup{font={footnotesize}}
\includegraphics[width =0.9\columnwidth]{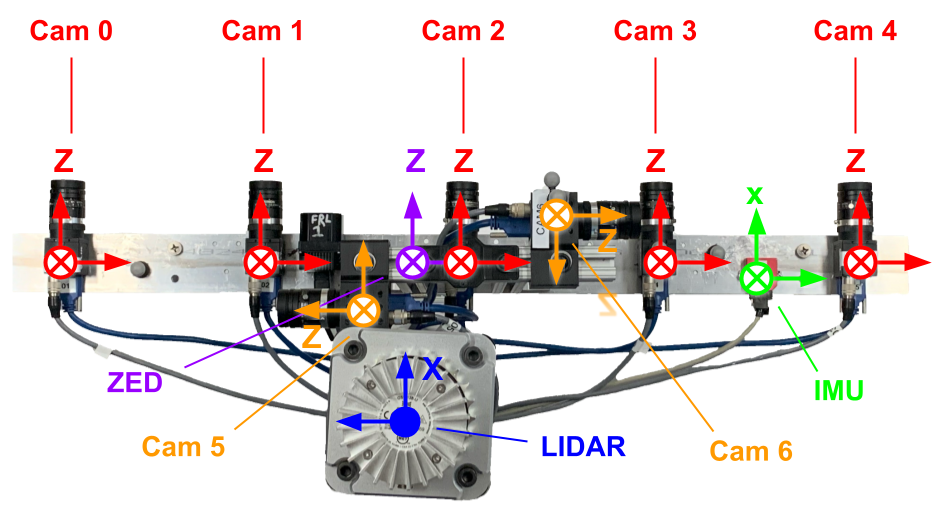}
\caption{Top view of the sensor rig showing sensor frames for the front-facing camera array (red), the non-overlapping side cameras (orange), the ZED camera (purple), the IMU (green) and the lidar (blue). Note the above image follows the convention that $\otimes$ indicates an axis into the plane of the image, and $\bullet$ indicates an axis out of the plane of the image. All of the cameras are z-axis forward, y-axis down.}
\label{fig:coord_frames}
\end{figure}

Our dataset described in table \ref{tab:dataset_sensors} consists of visual, inertial, and lidar sensor data, allowing for multi-modal SLAM evaluations. The specifications of each sensor are detailed in table \ref{tab:sensor_setup}. The entire sensor suite shown in figure \ref{fig:system} is time synchronized and spatially calibrated across all sensors for accurate data capture and analysis as shown in figure \ref{fig:coord_frames}.

To the best of our knowledge, this is the first dataset that has continuous multi-floor data for SLAM, and we know of no algorithm that is capable of processing the uninterrupted data across multiple floors into an accurate map of the entire building in an autonomous manner. The dataset presents new fundamental challenges to further the research on informing design decisions and algorithmic choices in performing SLAM with higher reliability. Even if (or when) this becomes possible, the dataset poses interesting questions related to localization due to symmetry across floors. This dataset serves to complement the recent multi-modal benchmarking datasets \cite{ramezani2020newer}\cite{helmberger2022hilti}\cite{jiang2021rellis}. The contributions of this paper can be summarized as:
\begin{itemize}
    \item It outlines challenging multi-modal indoor datasets covering a variety of scenarios including featureless spaces, reflective surfaces, and multi-storeyed sequences.
    \item The multi-storeyed sequences, as is typical with modern architecture, features floors that are essentially identical in design and layout which leads to perceptual aliasing scenarios. These scenarios trip up state-of-the-art SLAM algorithms, the vast majority of which rely on bag of words models for relocalization and loop closure.
    \item It features an extensive set of sensors consisting of seven cameras, a high-resolution lidar, and an IMU. All the sensors are hardware synchronized and calibrated across the entire sensor suite.
    \item We have benchmarked several state-of-the-art algorithms across the visual, visual-inertial, and lidar SLAM methodologies and present a comparison among these different algorithms and sensor modalities that highlights their individual strengths and areas where there are engineering or fundamental theoretical issues that the community may need to focus on. 
\end{itemize}


\section{Related Work}
Several SLAM datasets exist in the literature which vary in regards to the data acquisition environment, varying sensing modalities, type of motions, degree of difficulty, number of sensors, and synchronization of the data capture. The table \ref{tab:dataset_sensors} summarizes several multi-modal datasets closely related to us.

KITTI\cite{geiger2013vision} is one of the first and most popular benchmarking multi-modal datasets motivated by self-driving cars. It has a linear array of four cameras consisting of two stereo pairs - One RGB and one grayscale, a lidar, an IMU, and a GPS. Following this, many outdoor urban datasets emerged in the domain of autonomous driving, such as \cite{maddern20171}\cite{yogamani2019woodscape}\cite{agarwal2020ford}, which allowed the evaluation of various odometry and SLAM algorithms. 
Many earlier indoor SLAM datasets targeted visual odometry(VO) and visual-inertial odometry (VIO) tasks for monocular and stereo systems. The TUM\cite{schubert2018tum} and EUROC\cite{burri2016euroc} datasets are extensively used for benchmarking VO and VIO solutions. These datasets have global shutter stereo cameras, hardware synchronized with the IMU, and millimeter-accurate ground truth from motion capture systems. 

A few recent efforts \cite{ramezani2020newer}\cite{pfrommer2017penncosyvio} gathered multi-sensor (beyond stereo) and multi-modal data in urban indoor environments. PennCOSYVIO\cite{pfrommer2017penncosyvio}, is collected at Upenn's campus area with a stereo VI sensor, two project Tango devices, and three GoPro cameras arranged in a minimally overlapping configuration. The sensors are mounted on a handheld platform and carried across indoor and outdoor areas. The ground truth is provided using fiducial markers placed along the trajectories. The Newer College Dataset\cite{ramezani2020newer} and its extension contain synchronized image data from the Alphasense sensor with four cameras - two facing forward and two on the side as well as a lidar mounted on a handheld device. 

More recently, the Hilti\cite{helmberger2022hilti} and Hilti-Oxford\cite{zhang2022hilti} datasets attracted a lot of attention through their SLAM challenge, where multiple teams from both academia and industry participated. The main objective of this dataset is to push the limits of the state-of-the-art multi-sensor SLAM algorithms to aid real-world applications. There are indoor and outdoor sequences of construction sites and parking areas that contain some challenging scenarios of abrupt and fast motions and featureless areas. These datasets are collected with an Alphasense five-camera module with a stereo pair, three non-overlapping wide-angle cameras, an IMU, and two laser scanners.

All these datasets contain challenging sequences with changing lighting and texture, challenging structures such as staircases, and featureless spaces. Our dataset also consists of multi-modal data with cameras, lidar, and inertial measurements. In addition to featuring the challenging scenarios mentioned above, our dataset showcases symmetrical structures located on multiple floors which present unique challenges due to perceptual aliasing.  

\section{Data Collection System}
\subsection{Hardware Setup}
We built a rigid multi-sensor rig consisting of seven cameras, five facing forward and two facing sideways, an inertial measurement unit (IMU), a zed 2i sensor, and a lidar. The description and configuration of the sensors is shown in \cref{tab:sensor_setup}. The sensors' placement and coordinate frames are shown in the schematic figure \ref{fig:coord_frames}. The cameras are arranged to accommodate overlapping and non-overlapping configurations. The front-facing multi-stereo camera array, together with the left and right cameras, collectively yields a 171-degree field of view. We use Ouster's 128-beam high-resolution lidar, which gives high-density point clouds with 130,000 points. All the cameras except the ZED stereo cameras are hardware synchronized with IMU at 20 frames per second. We built a buffer circuit where the IMU sends a signal to trigger all the cameras simultaneously. The lidar and zed sensor are software/network time synchronized with the other sensor streams. All the sensor timestamps are assigned based on the hardware trigger in combination with the computer's system clock. The multi-sensor rig and a Dell XPS laptop with 32GB RAM were mounted on a Clearpath's Ridgeback robotic platform and driven using a joystick across multiple floors of two of the Northeastern University's buildings for data collection. The Zed 2i sensor is mounted for data collection in the Snell library building and is unavailable for the ISEC building.

\begin{table}[t!]
\captionsetup{font={footnotesize}}
\caption{Description of various sensors and their settings used to collect our dataset. Note that Zed2i sensor is available only in the Snell dataset.}
\label{tab:sensor_setup}
\resizebox{\columnwidth}{!}{%
\begin{tabular}{llll}
\hline
\textbf{Sensor} &
  \textbf{No} &
  \textbf{Type} &
  \textbf{Description} \\ \hline \hline
Camera &
  7 &
  \begin{tabular}[c]{@{}l@{}}FLIR \\ Blackfly S \\ USB3\end{tabular} &
  \begin{tabular}[c]{@{}l@{}}1.3 megapixel color cameras \\ with a resolution of 720 x 540 \\ and FoV of 57 $^\circ$ at 20 hz.\end{tabular} \\
Lidar &
  1 &
  \begin{tabular}[c]{@{}l@{}}Ouster\\ OS-128\end{tabular} &
  \begin{tabular}[c]{@{}l@{}}128 channel LiDAR with \\ vertical FoV of 45$^\circ$ \\ at 10 Hz\end{tabular} \\
\begin{tabular}[c]{@{}l@{}}RGB-D\\ camera\end{tabular} &
  1 &
  Zed 2i &
  \begin{tabular}[c]{@{}l@{}}stereo cameras with resolution of \\ 1280 x 720 at 15 Hz, \\ IMU at 200 Hz\end{tabular} \\
IMU &
  1 &
  \begin{tabular}[c]{@{}l@{}}Vectornav \\ 100\end{tabular} &
  9-DOF IMU running at 200 Hz. \\ \hline
\end{tabular}%
}

\end{table}

\subsection{Calibration}
We obtain both intrinsic and extrinsic calibration for cameras, IMU, and lidar by applying different methods. We used Kalibr\cite{furgale2013unified} to obtain the intrinsic and extrinsic parameters of the overlapping set of cameras and zed stereo cameras using a checkerboard target. For the side-facing cameras, it is not possible to use the same multi-camera calibration methods as we need the cameras to observe a single stationary target at the same time to find the correspondences and solve for the relative transformation. Instead, we perform IMU-camera calibration independently for the two side-facing cameras and the center front-facing camera to obtain $T_{c_i}^{IMU}$ and chain the camera-IMU transformations to get the inter-camera relative transforms using $T_{c_j}^{c_i} = (T_{c_i}^{IMU})^{-1} T_{c_j}^{IMU}$. We used target-based open source software packages \cite{Dhall2017LiDARCameraCU} and \cite{Zhou2018AutomaticEC} to obtain the lidar-camera extrinsic calibration parameters but noticed some misalignment of the point cloud with images which amplifies with range. We adjust the error by manually aligning the lidar point cloud with camera data. 


\subsection{Ground truth} 
 Ground truth poses are essential to test and evaluate the accuracy of the SLAM algorithms. However, generating ground truth trajectories in indoor environments is a challenging task due to the lack of GPS signals and range limitations of popular indoor ground truthing mechanisms like MOCAP. There are additional challenges particular to our dataset, where the robot moves across multiple floors which makes it impossible to deploy a MOCAP system to track the robot. 
 Given the necessity of ground-truth data for benchmarking novel algorithms, we used fiducial markers-based ground truth. These markers were used as stationary targets to localize the robot when they came into the cameras' field of view. We carefully placed multiple fiducial markers made of April tags\cite{apriltag} near the elevators on each floor. The location is chosen so as to allow the April tags to be visible at the start and end of trajectories on each floor as we drive the robot in loops, as well as at the transits across floors when we enter and exit the elevators. We explain how we compute the error metrics in detail in section (V-A).

\newcommand{\datawidth}{0.18\textwidth}
\begin{figure*}[t!]
\centering
\captionsetup{font={footnotesize}}
\begin{tabular}{c}
\subfloat[]{\includegraphics[width =\datawidth]{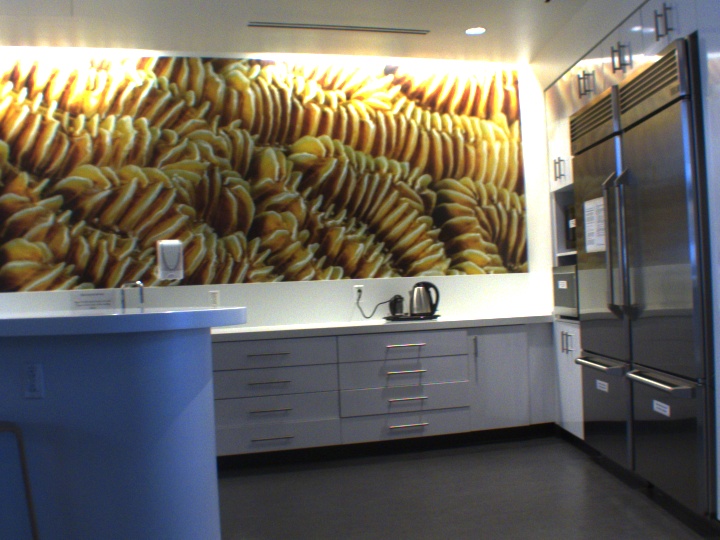}}
\hspace{1mm}
\subfloat[]{\includegraphics[width =\datawidth]{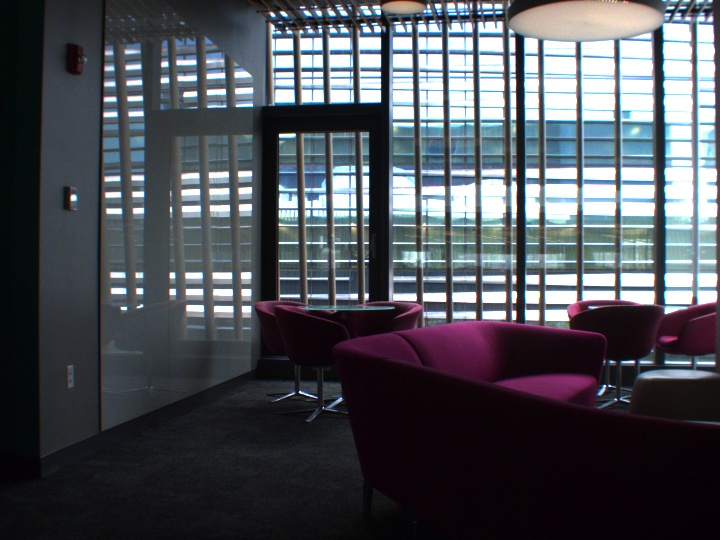}}
\hspace{1mm}
\subfloat[]{\includegraphics[width =\datawidth]{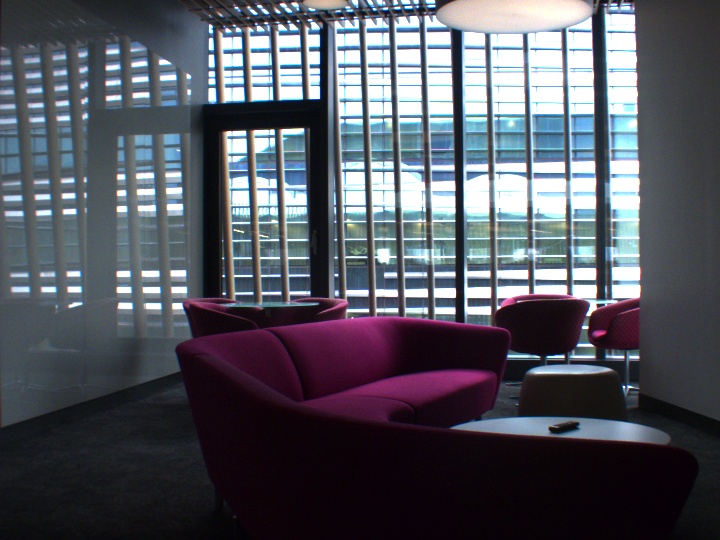}}
\hspace{1mm}
\subfloat[]{\includegraphics[width =\datawidth]{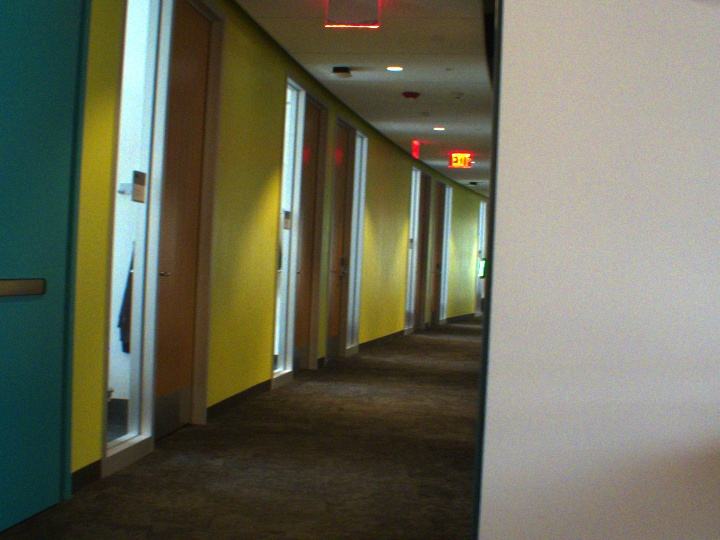}}
\hspace{1mm}
\subfloat[]{\includegraphics[width =\datawidth]{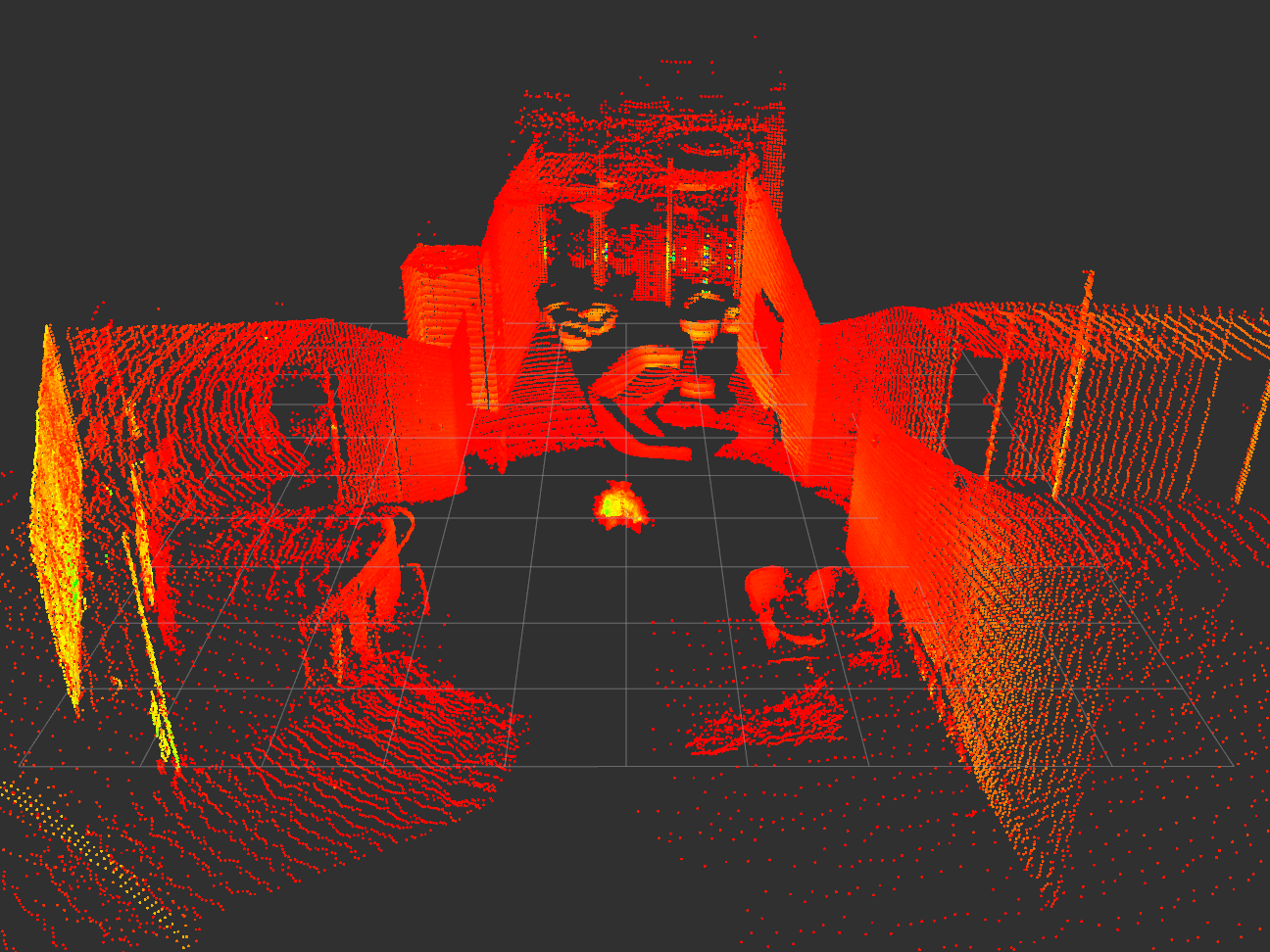}}
\end{tabular}
\caption{This figure shows a sample of the various available data streams, showing (a) the left facing side camera (Cam5), (b) and (c) a stereo pair from the front facing array (Cam1 \& Cam3), (d) the right facing side camera (Cam6), and (e) the lidar point cloud.}
\label{fig:data_streams}
\end{figure*}

\section{Dataset}

\begin{figure}[t!]
\centering
\captionsetup{font={footnotesize}}
\includegraphics[width =0.9\columnwidth]{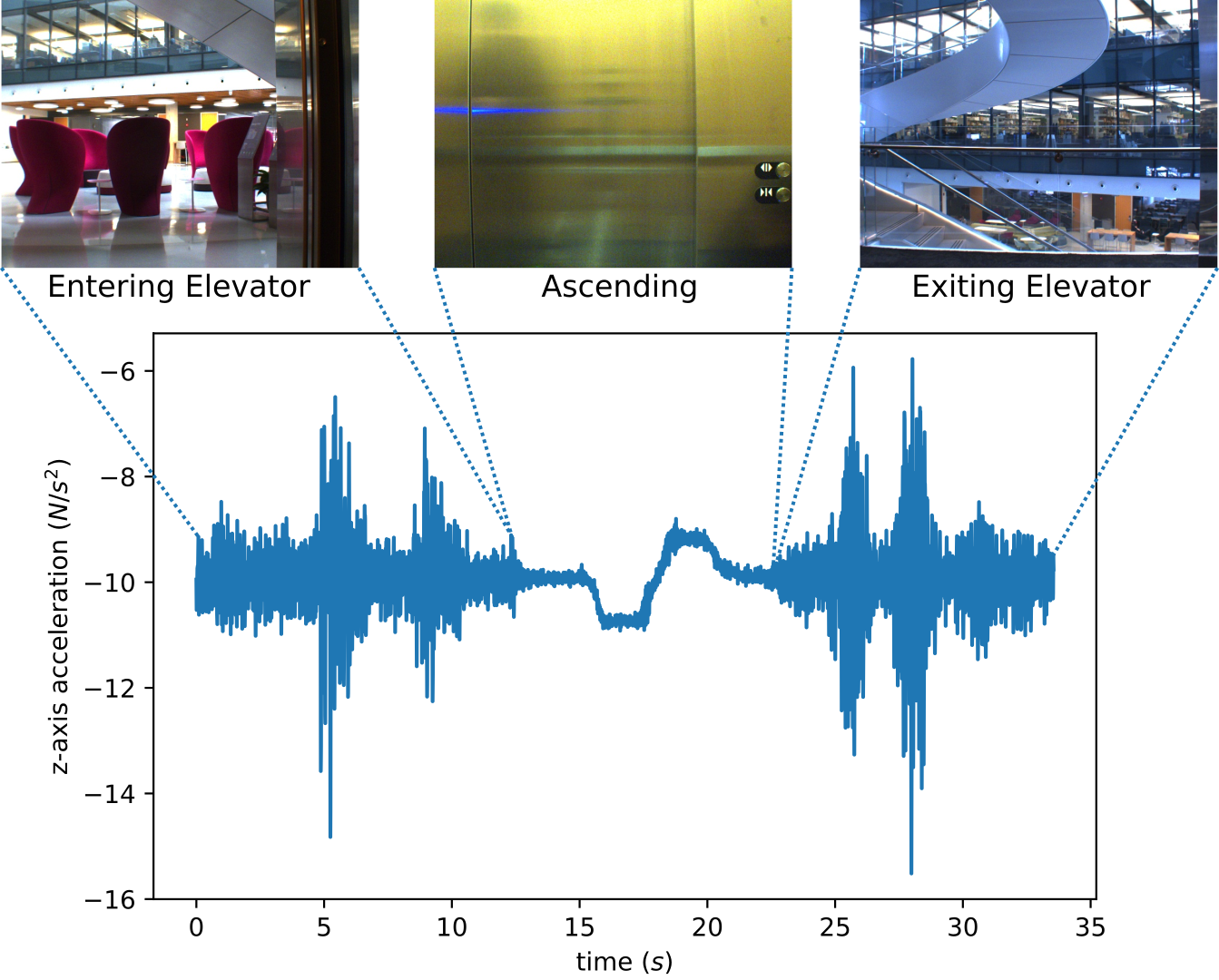}
\caption{The full dataset has several points where the robot enters an elevator. The vision-only and lidar SLAM algorithms are not able to handle a scenario where significant movement is not rendered in the data. This figure shows the z-axis IMU acceleration as the robot ascends in the elevator from the first to the second floor. The spikes as the robot enters and exits the elevator correspond to the robot wheels rolling over the gap between the elevator and the hallway.}
\label{fig:elevator_data}
\end{figure}

We collected two large datasets of indoor office environments. The datasets were generated by driving in a loop through different floors of Northeastern University's campus buildings and traveling by elevator between floors. The trajectories include several challenging scenarios that occur on a day-to-day basis, including narrow corridors, featureless spaces, jerky and fast motions, sudden turns, and dynamic objects, which are commonly encountered by a mobile robot in urban environments. All the trajectories have loops to allow the SLAM systems to perform loop closure and compute drift when continuous ground truth is unavailable. All the data is collected using ROS drivers for the respective sensors. The dataset details, including location, length of the trajectory, and ground truth, are consolidated in the \cref{tab:dataset_desc}. 

\subsection{ISEC Dataset:}
The multiple-floor trajectory was collected in Northeastern University's Interdisciplinary Science and Engineering Complex (ISEC) building. There are four complete floor sequences in the dataset and multiple transit sequences, which include five elevator rides between floors. We start on the $5^{th}$ floor and drive through the space such that it contains two loops, where the second part is a trajectory down and back a long corridor with a loop closure. We then take an elevator ride to the $1^{st}$ floor, where we acquire another loop. The $1^{st}$ floor sequence contains more dynamic objects, glass, and distinct architecture when compared to the other floors. From the $1^{st}$ floor, we transit through a long corridor with white walls, take an elevator to $3^{rd}$ floor, and then another one to the $4^{th}$ floor. We cover the $4^{th}$ floor and then proceed to the $2^{nd}$ floor before taking the final elevator ride to the $5^{th}$ floor, where we started. The first loop of $5^{th}$ floor, $4^{th}$, and $2^{nd}$ floor sequences are nearly identical with similar-looking office spaces. Thus, these indoor sequences cover areas with good and bad natural lighting, a mix of artificial and natural light, reflections, and dynamic content, such as students. The indoor data snapshots are shown in the figure \ref{fig:data_streams}. 

\subsection{Snell Library Dataset}
This dataset is collected across multiple floors of Northeastern University's library building by taking elevator rides similar to the ISEC dataset. In general, the Snell dataset has better visual features but has longer trajectories with more dynamic content than the ISEC dataset, which can be a failure point for SLAM algorithms. This sequence poses a challenge to SLAM algorithms to map highly dynamic environments. We travel through 3 floors of the building with loop closures on each floor and where the $1^{st}$ floor's appearance differs from the other two floors.
\begin{table*}[ht!]
\centering
\scriptsize
\captionsetup{font={small}}
\caption{A comprehensive list of all the sequences in our dataset and their description. Trajectory lengths are approximate and should only be used for qualitative comparison. They were derived from the best available trajectory estimate for each segment. This was typically lidar odometry (LegoLOAM) for the loop sequences and VIO (Basalt or VINS Fusion) for sequences inside elevators. See section V for more details on trajectory estimates.}
\label{tab:dataset_desc}
\resizebox{0.9\textwidth}{!}{%
\begin{tabular}{|llllllll|}
\hline
\multicolumn{8}{|c|}{\textbf{Datasets}} \\ \hline
\multicolumn{1}{|l|}{\textbf{Label}} &
  \multicolumn{1}{l|}{\textbf{Size (GB)}} &
  \multicolumn{1}{p{1cm}|}{\textbf{Duration (s)}} &
  \multicolumn{1}{p{1cm}|}{\textbf{Appx. Length (m)}} &
  \multicolumn{4}{c|}{\textbf{Description}} \\ \hline
\multicolumn{8}{|c|}{\textbf{ISEC}} \\ \hline
\multicolumn{1}{|c|}{full\_sequence} &
  \multicolumn{1}{l|}{515.0} &
  \multicolumn{1}{l|}{1539.70} &
  \multicolumn{1}{l|}{782} &
  \multicolumn{4}{p{5cm}|}{reflective surfaces, minimal dynamic content, daylight, symmetric floors, elevators, open atrium} \\ \hline

\multicolumn{1}{|l|}{5th\_floor} &
  \multicolumn{1}{l|}{145.8} &
  \multicolumn{1}{l|}{437.86} &
  \multicolumn{1}{l|}{187} &
  \multicolumn{4}{p{5cm}|}{one loop, one out and back} \\ \hline
  
\multicolumn{1}{|l|}{transit\_5\_to\_1} &
  \multicolumn{1}{l|}{36.8} &
  \multicolumn{1}{l|}{109.00} &
  \multicolumn{1}{l|}{*} &
  \multicolumn{1}{p{5cm}|}{transit from 5th to 1st floor in middle elevator} \\ \hline
  
\multicolumn{1}{|l|}{1st\_floor} &
  \multicolumn{1}{l|}{43.0} &
  \multicolumn{1}{l|}{125.58} &
  \multicolumn{1}{l|}{65} &
  \multicolumn{1}{p{5cm}|}{one loop, open layout different from other floors, many exterior windows} \\ \hline

\multicolumn{1}{|l|}{transit\_1\_to\_4} &
  \multicolumn{1}{l|}{112.4} &
  \multicolumn{1}{l|}{337.40} &
  \multicolumn{1}{l|}{144} &
  \multicolumn{1}{p{5cm}|}{transit across 1st floor, up to 3rd floor in freight elevator, across 3rd floor, up to 4th floor in right elevator} \\ \hline
  
\multicolumn{1}{|l|}{4th\_floor} &
  \multicolumn{1}{l|}{43.2} &
  \multicolumn{1}{l|}{131.00} &
  \multicolumn{1}{l|}{66} &
  \multicolumn{1}{p{5cm}|}{one loop, some dynamic content towards end} \\ \hline
  
\multicolumn{1}{|l|}{transit\_4\_to\_2} &
  \multicolumn{1}{l|}{21.9} &
  \multicolumn{1}{l|}{65.00} &
  \multicolumn{1}{l|}{22} &
  \multicolumn{1}{p{5cm}|}{transit from 4th floor to second floor in right elevator, } \\ \hline
  
\multicolumn{1}{|l|}{2nd\_floor} &
  \multicolumn{1}{l|}{89.7} &
  \multicolumn{1}{l|}{266.00} &
  \multicolumn{1}{l|}{128} &
  \multicolumn{1}{p{5cm}|}{two loops in a figure eight} \\ \hline

\multicolumn{1}{|l|}{transit\_2\_to\_5} &
  \multicolumn{1}{l|}{22.2} &
  \multicolumn{1}{l|}{65.86} &
  \multicolumn{1}{l|}{128} &
  \multicolumn{1}{p{5cm}|}{transit from 2nd floor to fifth floor in right elevator} \\ \hline
\multicolumn{8}{|c|}{\textbf{SNELL LIBRARY}} \\ \hline
\multicolumn{1}{|l|}{full\_sequence} &
  \multicolumn{1}{l|}{573.5} &
  \multicolumn{1}{l|}{1,700.6} &
  \multicolumn{1}{l|}{699} &
  \multicolumn{1}{p{5cm}|}{feature rich rooms, featureless hallways, many obstacles, stationary and dynamic people in scene} \\ \hline

\multicolumn{1}{|l|}{1st\_floor} &
  \multicolumn{1}{l|}{144.6} &
  \multicolumn{1}{l|}{428.70} &
  \multicolumn{1}{l|}{221} &
  \multicolumn{1}{p{5cm}|}{two loops with shared segment, some dynamic content} \\ \hline

\multicolumn{1}{|l|}{transit\_1\_to\_3} &
  \multicolumn{1}{l|}{28.3} &
  \multicolumn{1}{l|}{84.00} &
  \multicolumn{1}{l|}{*} &
  \multicolumn{1}{p{5cm}|}{transit from 1st floor to 3rd floor in left elevator} \\ \hline

\multicolumn{1}{|l|}{3rd\_floor} &
  \multicolumn{1}{l|}{213.7} &
  \multicolumn{1}{l|}{633.59} &
  \multicolumn{1}{l|}{345} &
  \multicolumn{1}{p{5cm}|}{two concentric loops with two shared segments, narrow corridor with dynamic content, near field obstructions} \\ \hline

\multicolumn{1}{|l|}{transit\_3\_to\_2} &
  \multicolumn{1}{l|}{27.8} &
  \multicolumn{1}{l|}{82.41} &
  \multicolumn{1}{l|}{*} &
  \multicolumn{1}{p{5cm}|}{transit from 3rd floor to 2nd floor in right elevator} \\ \hline

\multicolumn{1}{|l|}{2nd\_floor} &
  \multicolumn{1}{l|}{126.1} &
  \multicolumn{1}{l|}{374.00} &
  \multicolumn{1}{l|}{186} &
  \multicolumn{1}{p{5cm}|}{one loop, out and back in featureless corridor} \\ \hline

\multicolumn{1}{|l|}{transit\_2\_to\_1} &
  \multicolumn{1}{l|}{33.0} &
  \multicolumn{1}{l|}{97.90} &
  \multicolumn{1}{l|}{*} &
  \multicolumn{1}{p{5cm}|}{transit from 2nd floor to 1st floor in right elevator, dynamic objects cover FOV near end} \\ \hline

\end{tabular}%
}

\end{table*}

\section{Benchmarking the SOTA}

To demonstrate the quality and usefulness of the dataset, we benchmark across a set of well-known state-of-the-art SLAM algorithms. The investigated algorithms are selected so as to have a broad coverage of the field, including visual SLAM, visual-inertial, and lidar-based solutions. The complete list of algorithms can be seen in \cref{tab:benchmarking_table}. We also provide the configuration settings we use to run each algorithm.

\subsection{Evaluation}
We run the visual and visual-inertial methods in stereo mode to use the metric scale in the evaluation. We use front-facing cameras 1 and 3 as the stereo pair for each algorithm (see figure \ref{fig:coord_frames} for camera placement), except for MCSLAM, which uses the full array of front-facing cameras. This pair was selected as a compromise between a wider stereo baseline and camera proximity to the IMU. We evaluate visual SLAM algorithms on trajectories collected on each floor, whereas visual-inertial algorithms are also evaluated during the transit sequences in the elevators. The elevator sequences are particularly valuable as they give us an insight into the utility of the inertial sensors when vision is ineffective, which is discussed in section V-B. We conducted quantitative analysis on the ISEC dataset by computing error metrics and limited the Snell Library dataset to qualitative results.

In most portions of the dataset, lidar odometry computed using Lego-LOAM can be used as a reasonable ground truth, but it does fail in some portions, and while the results are very good qualitatively, it is non-trivial to compute an upper bound on trajectory errors in the resulting pseudo ground truth. To avoid this kind of analysis, we provide a more limited ground truth evaluation for the dataset using fiducial markers, with a separate evaluation for each floor. We mount an AprilTag \cite{apriltag} tracking target on walls that are visible at the beginning and end of the trajectory at each floor, giving a fixed reference point from which to compute the drift accumulated by each algorithm. For the initial and final portions, when the target is visible, we compute the ground truth poses of the robot $\mathbf{T_{rig}^{target}}$ by localizing it with respect to the target using PnP ransac\cite{hartley2003multiple} followed least squares optimization. To align the trajectories, we estimate the rigid body transformation $\mathbf{T_O^{target}} \in \mathbf{se(3)}$ using the positions $\mathbf{t_{rig}^{O^{(i)}}}$ and $\mathbf{t_{rig}^{target^{(i)}}}$ of the tracked and ground truth poses belonging to the starting segment of the trajectory such that
\begin{equation}
    T_O^{target} = \underset{T_O^{target}t_{rig}}{argmin} \;\;  \sum_i\|T_O^{target}t_{rig}^{O^{(i)}} - t_{rig}^{target^{(i)}} \|^2 
\end{equation}
\begin{table*}[t!]
\centering
\captionsetup{font={small}}
\caption{This table outlines the performance of various algorithms on the ISEC dataset. We evaluate each algorithm on loops on the $5^{th}$, $1^{st}$, $4^{th}$, and $2^{nd}$ floors, in the order they appear in the continuous dataset. Inertial algorithms are also evaluated on the full dataset, which includes elevator transits between floors. Results are reported as the absolute transnational error at the final position in meters, and as a percentage of the estimated trajectory length.We run each algorithm with loop closure disabled, because most algorithms can use the AprilTag markers to form a loop closure, bringing the error close to zero, which does not produce a useful performance metric. While testing the $2^{nd}$ and $4^{th}$ floors individually, vins-Fusion resulted in unusually high drift and was left out of this analysis. All vins algorithms surprisingly display higher drift than the visual counterparts due to issues with initialization. We perform a loop closure analysis in Discussion subsection A.}
\label{tab:benchmarking_table}
\begin{tabular}{|lcccccccccc|}
\hline
\multicolumn{1}{|l|}{\multirow{2}{*}{Algorithm}}      & \multicolumn{2}{c|}{5\textsuperscript{th} Floor}          & \multicolumn{2}{c|}{1\textsuperscript{st} Floor}          & \multicolumn{2}{c|}{4\textsuperscript{th} Floor}           & \multicolumn{2}{c|}{2\textsuperscript{nd} Floor}               & \multicolumn{2}{c|}{Full Dataset}    \\ \cline{2-11} 
\multicolumn{1}{|l|}{}                                & \multicolumn{1}{c|}{ATE(m)} & \multicolumn{1}{c|}{\%}     & \multicolumn{1}{c|}{ATE(m)} & \multicolumn{1}{c|}{\%}     & \multicolumn{1}{c|}{ATE(m)} & \multicolumn{1}{c|}{\%}      & \multicolumn{1}{c|}{ATE(m)}    & \multicolumn{1}{c|}{\%}       & \multicolumn{1}{c|}{ATE(m)} & \%     \\ \hline
\multicolumn{11}{|c|}{Visual SLAM}                                                                                                                                                                                                                                                                                                                 \\ \hline
\multicolumn{1}{|l|}{ORB-SLAM3\cite{ORBSLAM3_TRO}}    & \multicolumn{1}{c|}{0.516}  & \multicolumn{1}{c|}{0.28\%} & \multicolumn{1}{c|}{0.949}  & \multicolumn{1}{c|}{1.46\%} & \multicolumn{1}{c|}{0.483}  & \multicolumn{1}{c|}{0.73\%}  & \multicolumn{1}{c|}{0.310}     & \multicolumn{1}{c|}{0.24\%}   & \multicolumn{1}{c|}{--}     & --     \\ \hline
\multicolumn{1}{|l|}{SVO\cite{Forster17troSVO}}       & \multicolumn{1}{c|}{0.626}  & \multicolumn{1}{c|}{0.33\%} & \multicolumn{1}{c|}{0.720}  & \multicolumn{1}{c|}{1.11\%} & \multicolumn{1}{c|}{0.482}  & \multicolumn{1}{c|}{0.73\%}  & \multicolumn{1}{c|}{0.371}     & \multicolumn{1}{c|}{0.29\%}   & \multicolumn{1}{c|}{--}     & --     \\ \hline
\multicolumn{1}{|l|}{MCSLAM}                          & \multicolumn{1}{c|}{0.778}  & \multicolumn{1}{c|}{0.42\%} & \multicolumn{1}{c|}{1.085}  & \multicolumn{1}{c|}{1.67\%} & \multicolumn{1}{c|}{0.484}  & \multicolumn{1}{c|}{0.73\%}  & \multicolumn{1}{c|}{0.458}     & \multicolumn{1}{c|}{0.36\%}   & \multicolumn{1}{c|}{--}     & --     \\ \hline
\multicolumn{11}{|c|}{Visual-Inertial}                                                                                                                                                                                                                                                                                                             \\ \hline
\multicolumn{1}{|l|}{vins-Fusion\cite{qin2018online}} & \multicolumn{1}{c|}{1.120}  & \multicolumn{1}{c|}{0.60\%} & \multicolumn{1}{c|}{2.265}  & \multicolumn{1}{c|}{3.48\%} & \multicolumn{1}{c|}{-} & \multicolumn{1}{c|}{-} & \multicolumn{1}{c|}{-} & \multicolumn{1}{c|}{-} & \multicolumn{1}{c|}{15.844} & 2.03\% \\ \hline
\multicolumn{1}{|l|}{Basalt\cite{usenko19nfr}}        & \multicolumn{1}{c|}{1.214}  & \multicolumn{1}{c|}{0.65\%} & \multicolumn{1}{c|}{4.043}  & \multicolumn{1}{c|}{6.22\%} & \multicolumn{1}{c|}{1.809}  & \multicolumn{1}{c|}{2.74\%}  & \multicolumn{1}{c|}{3.054}     & \multicolumn{1}{c|}{2.39\%}   & \multicolumn{1}{c|}{1.753}  & 0.22\% \\ \hline
\multicolumn{1}{|l|}{SVO-inertial}                    & \multicolumn{1}{c|}{0.649}  & \multicolumn{1}{c|}{0.35\%} & \multicolumn{1}{c|}{2.447}  & \multicolumn{1}{c|}{3.76\%} & \multicolumn{1}{c|}{0.558}  & \multicolumn{1}{c|}{0.85\%}  & \multicolumn{1}{c|}{0.621}     & \multicolumn{1}{c|}{0.48\%}   & \multicolumn{1}{c|}{16.202} & 2.07\% \\ \hline
\multicolumn{11}{|c|}{Deep Learning}                                                                                                                                                                                                                                                                                                               \\ \hline
\multicolumn{1}{|l|}{Droid SLAM\cite{teed2021droid}}  & \multicolumn{1}{c|}{0.441}  & \multicolumn{1}{c|}{0.24\%} & \multicolumn{1}{c|}{0.666}  & \multicolumn{1}{c|}{1.02\%} & \multicolumn{1}{c|}{0.112}  & \multicolumn{1}{c|}{0.17\%}  & \multicolumn{1}{c|}{0.214}     & \multicolumn{1}{c|}{0.17\%}   & \multicolumn{1}{c|}{--}     & --     \\ \hline
\multicolumn{11}{|c|}{LIDAR}                                                                                                                                                                                                                                                                                                                       \\ \hline
\multicolumn{1}{|l|}{LEGO LOAM\cite{legoloam2018}}    & \multicolumn{1}{c|}{0.395}  & \multicolumn{1}{c|}{0.21\%} & \multicolumn{1}{c|}{0.256}  & \multicolumn{1}{c|}{0.39\%} & \multicolumn{1}{c|}{0.789}  & \multicolumn{1}{c|}{1.20\%}  & \multicolumn{1}{c|}{0.286}     & \multicolumn{1}{c|}{0.22\%}   & \multicolumn{1}{c|}{--}     & --     \\ \hline
\end{tabular}
\vspace{-3mm}
\end{table*}

\newcommand{\chammelgewidth}{0.23\textwidth}
\begin{figure}[!]
\centering
\captionsetup{font={footnotesize}}
\subfloat[Glass Surfaces]{\includegraphics[width =\chammelgewidth]{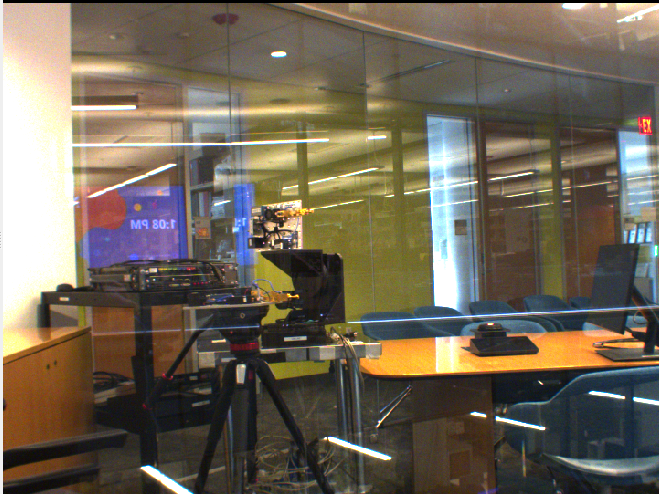}}
\hspace{1mm}
\subfloat[Elevator Areas]{\includegraphics[width =\chammelgewidth]{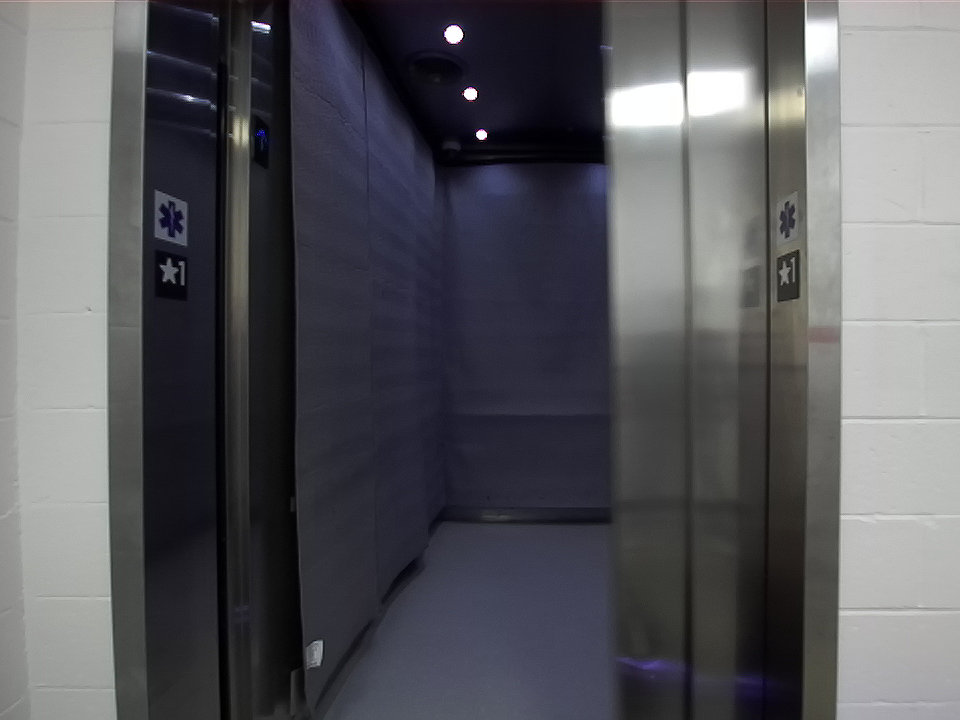}}
\hspace{1mm}
\caption{A sample of some challenging points in the dataset. Image (a) shows a glass wall with reflections that can introduce spurious features. Image (b) shows one of the elevator areas, where once the robot enters, the exteroceptive sensors such as LiDARs and cameras are fundamentally limited to track motion.}
\label{fig:challenge_areas}
\end{figure}

Once we have the transformation $\mathbf{T_O^{target}}$, we compute the total translational error or the drift at the end of the trajectory between the investigated algorithm's reported pose and the ground truth pose computed using the fixed markers. We report this final drift error for each investigated algorithm as the Absolute Translational Error(ATE), and as a percentage of the approximate total length of the trajectory in \cref{tab:benchmarking_table}. We compute this drift for each floor individually for all the algorithms. We compute the average of the ATEs accumulated at the April tags on different floors for inertial algorithms. The April tags are placed at exactly known locations on each floor so that they are displaced vertically by a fixed distance, which is verified from the building floor plan. 

\begin{figure*}[]
\centering
\captionsetup{font={footnotesize}}
\begin{tabular}{c}
\subfloat[]{\includegraphics[width =0.45\textwidth]{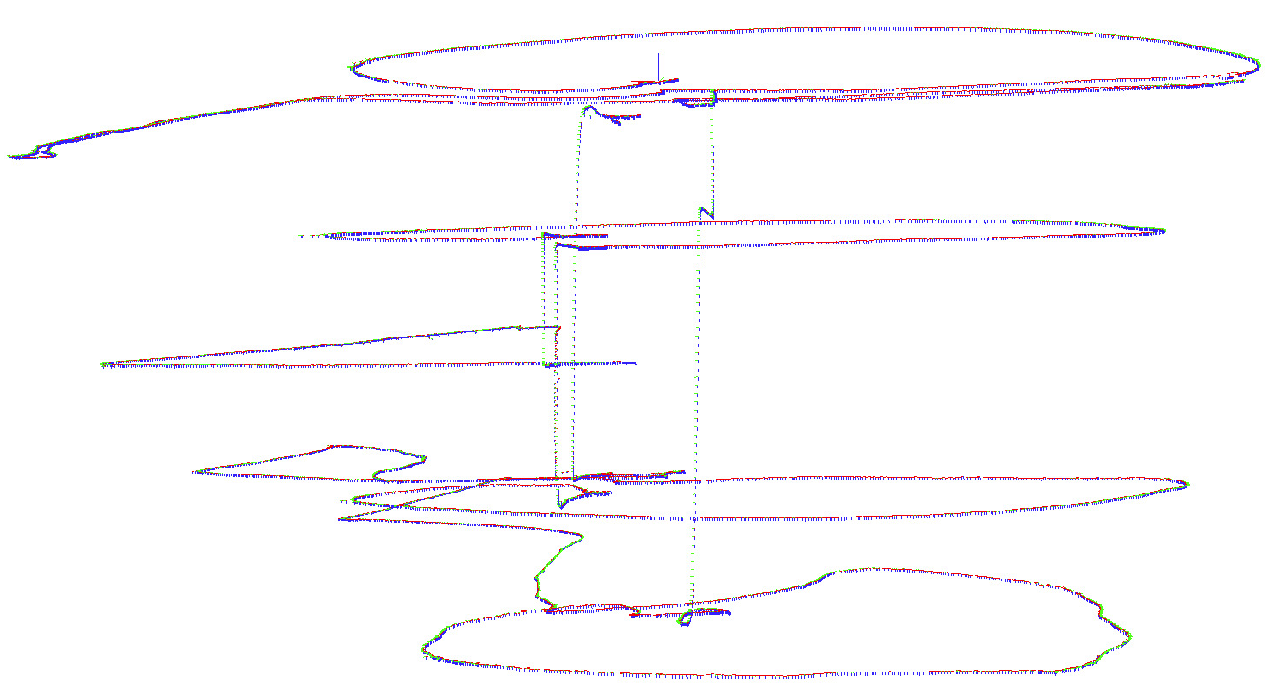}}
\subfloat[]{\includegraphics[width =0.45\textwidth]{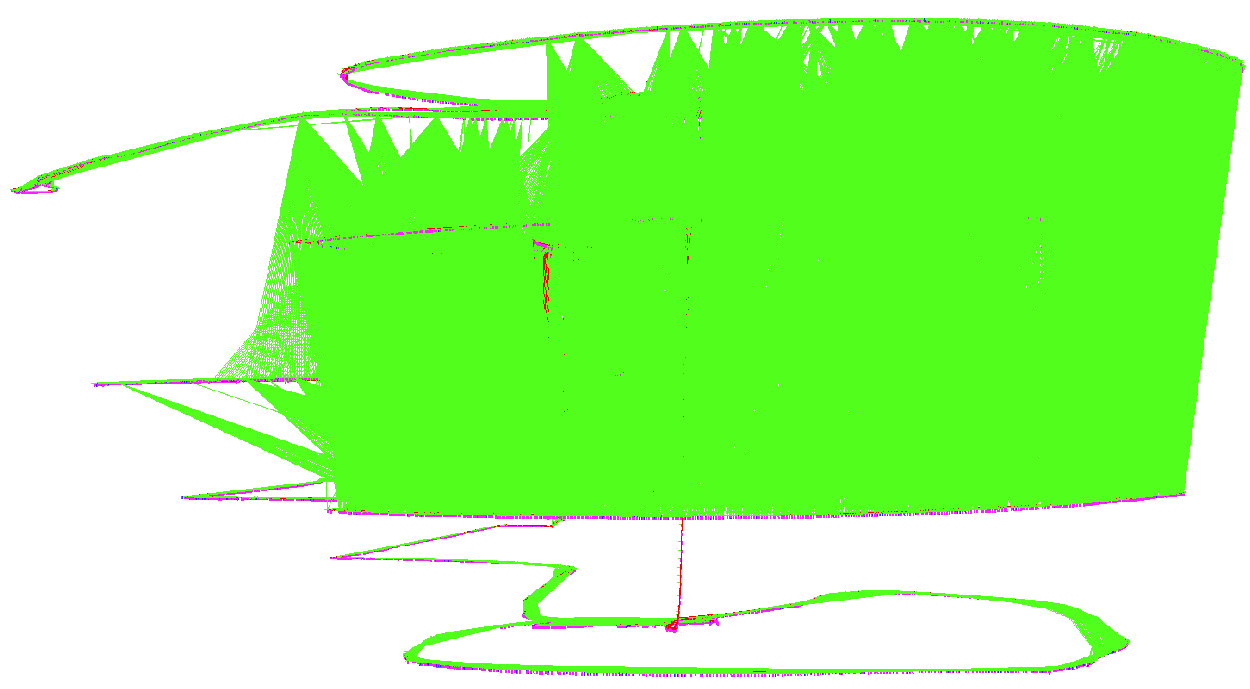}}
\end{tabular}
\caption{This shows the perceptual aliasing problem that is typical of modern buildings. (a) It shows the estimated trajectory of Basalt, a visual-inertial SLAM system for the full multi-floor sequence of the ISEC building without the loop detection, and (b) shows the same sequence run after the loop closure detection. Here the green line segments connecting floors are the incorrectly identified loop closure constraints between poses due to the similarity in appearance.}
\label{fig:perceptual_aliasing}
\end{figure*}

\begin{figure}
\centering
\captionsetup{font={footnotesize}}
\includegraphics[width =1.0\columnwidth]{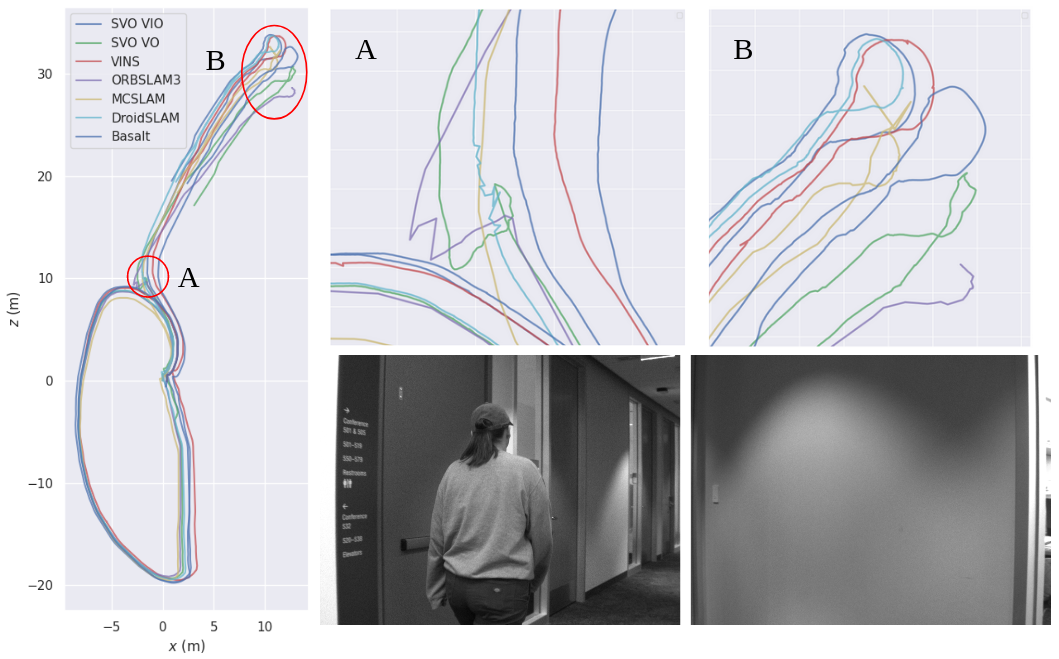}
\caption{This figure shows the 5th-floor trajectories calculated by the various algorithms with two highlighted areas, \textbf{(A)} and \textbf{(B)}. \textbf{(A)} shows a portion of the sequence with dynamic content and its impact on the trajectory estimates, resulting in jagged artifacts for the vision-only algorithms. \textbf{(B)} highlights a featureless environment during a tight turn, which caused incorrect trajectory estimates or failure in the vision-only algorithms.}
\label{fig:failure_points}
\end{figure}

\begin{figure}
\centering
\captionsetup{font={footnotesize}}
\includegraphics[width =1.0\columnwidth]{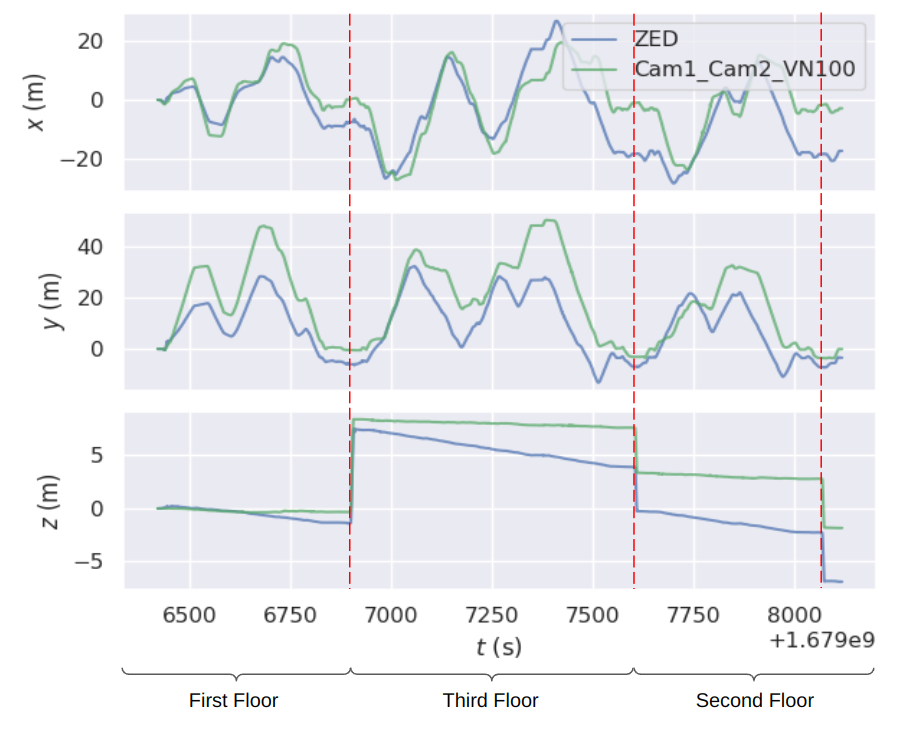}
\caption{This figure shows the difference in performance when we run VINS-Fusion on the Library dataset with the VN 100 IMU and ZED IMU. The figure compares x, y, z positions estimated by VINS fusion in both configurations. The red dotted lines show when we enter the elevator. On every floor, we come back to the starting position, and after 2nd floor, we come back to the first floor starting position again which was our origin. The figure clearly shows that VINS Fusion accumulates more drift with ZED sensor setup in all three axes.}
\label{fig:drift_library}
\end{figure}

\subsection{Discussion}
We want to point out that the accuracy metric does not fully describe the performance of a SLAM system. Evaluating drift from the beginning to the end of a trajectory can overlook essential details but is somewhat reflective of the pass-fail nature of real-world scenarios. A more comprehensive evaluation should look at features, like loop detection and closure, tracking failures, and map correction while considering reliability and robustness. In this section, we provide some qualitative assessments of the tested algorithms.

\subsubsection{Perceptual Aliasing}
Our dataset targets this primary challenge by showcasing multi-floor trajectories with similar-looking areas. Most of the vision-based SLAM frontends use a bag of words model \cite{GalvezTRO12} to compute the appearance-based similarity between images for loop detection. In addition, vision-only SLAM methods inherently lack the ability to recognize elevator motion. Based on the end-to-end runs of the algorithms, we observed that all the evaluated VO and VIO algorithms are prone to wrong loop closures, confusing one floor with another. This happens with the $5^{th}$, $4^{th}$, and $2^{nd}$ floors, which are symmetrical in structure, color, and layout.  This leads to incorrect loop constraints between poses belonging to different floors causing the entire trajectory of one floor to shift in space. Figure \ref{fig:perceptual_aliasing} shows the constraints as edges between the $2^{nd},3^{rd},4^{th}, \text{and } 5^{th}$ floors even though there is no direct visibility across them whereas the first floor remains disconnected since it is unique in appearance. As a result, the trajectories appear to be on the same floor, and the possibility of wrong loop detections is high. In the case of VIO, despite having a good sense that we are not on the same floor, incorrect loop detections still happen.

\subsubsection{Visual Degradation} 
Visual degradation occurs at multiple places along the trajectories when we encounter featureless spaces, reflective surfaces, and dynamic content, as shown in figure \ref{fig:challenge_areas}. All the algorithms run without tracking loss on the $1^{st}$, $2^{nd}$, and $4^{th}$ floors with minimal drift. In the presence of dynamic objects such as moving people, stereo visual slam algorithms cause jagged artifacts due to corrupted relative motion estimates, as shown in figure \ref{fig:failure_points}. Visual-inertial and multi-camera SLAM systems do not display these problems. The vision-only algorithms do not always provide accurate estimates when we run into plain walls, glass surfaces, and during elevator rides. Among vision-only methods, feature-based methods such as ORBSLAM3 are more prone to tracking failures when featureless walls are encountered. Direct methods like SVO can still track due to optical flow but result in incorrect pose estimates causing drift in the subsequent poses. DroidSLAM, which is a learning-based stereo method, also copes in featureless scenarios; however, it lacks scale. These problems are highlighted in figure \ref{fig:failure_points}.
Visual-inertial algorithms perform well in these scenarios due to the presence of a proprioceptive inertial sensing component, which can detect the physical motion of the vehicle. However, we observed that inertial sensing is not always effective when vision fails. For instance, when we ride in the elevator, the visual features detected on the elevator walls interfere with the inertial sensing leading to erroneous poses.

\subsubsection{Other Issues}
We have noticed that the performance of VIO algorithms heavily relies on the initial conditions and parameter tuning. In some sequences, the algorithms perform poorly when we start from specific points. Even starting the data with a time difference of +/- 2 seconds shifts the final drift by about 5 meters. The current SLAM algorithms have a massive list of different parameters that need to be tuned specifically to the dataset. These parameters are generally not standardized or consistent across different algorithms, and tuning them can be arduous. Learning-based methods have an edge in this regard since they do not need as much manual intervention. Additionally, the type and configuration of sensors also impact the performance of the algorithms, which is essential but is unfortunately one of the less researched topics. To demonstrate this, we compare the estimated trajectories of one of the visual-inertial algorithms (VINS-Fusion) executed on the ZED's stereo inertial system and the stereo configuration with VN100 IMU used in earlier evaluations on the complete multi-floor sequence of the Snell Library dataset. We observe that the two runs differ significantly, as shown in figure \ref{fig:drift_library}.   
 
\subsubsection{Potential usage}
The previous discussion clearly shows that the current algorithms fall short in performing large-scale indoor SLAM. There is much room for improvement in the various real-world scenarios discussed above. An upcoming research direction in this regard is to incorporate semantic information from vision into the SLAM framework, as explored in many recent works. One possible solution to improve loop closure detection would be to use contextual information specific to the location, structure, and objects to distinguish between the floors. There is also a need for better modeling of IMU data that captures the noise properties better \cite{nir2021high}, contributing to better SLAM back-ends.

\section{Conclusion}
We have presented a novel multi-modal SLAM dataset that contains visual, inertial, and lidar. The dataset contains several challenging sequences collected by driving a mobile robot across multiple floors of an open-concept office space with narrow corridors, featureless spaces, glass surfaces, and dynamic objects, which challenge the SLAM algorithms. One of the exciting features of our dataset is the symmetric and visually similar locations across different floors that cause perceptual aliasing. We evaluated several SLAM and visual odometry methods across different sensor modalities. The results demonstrate the limitations and areas of improvement in the current SOTA. The main goal of this dataset is to enable the development and testing of novel algorithms for indoor SLAM to address the various challenges discussed. We intend to expand the dataset to outdoors and add more challenging sequences in the future.

{\small
\bibliographystyle{IEEEbib}
\bibliography{references}
}
\end{document}